\documentclass[sigconf]{acmart}
\setcopyright{none}
\renewcommand\footnotetextcopyrightpermission[1]{\unskip}
\acmConference[WWW 2027]{The Web Conference 2027}{May 10--14, 2027}{Dublin, Ireland}

\usepackage{amsmath}
\usepackage{amsthm}
\usepackage{booktabs}
\usepackage{graphicx}
\usepackage{cleveref}
\usepackage{algorithm}
\usepackage[noend]{algorithmic}
\usepackage{xcolor}
\usepackage{balance} 

\begin{document}

\title{Text as an Index for Zero-Shot Cross-Domain Knowledge Tracing}


\author{Yin Yi}
\affiliation{
  \department{Institutes of Physical Science and Information Technology}
  \institution{Anhui University}
  \city{Hefei}
  \country{China}
}

\author{Haiping Ma*}
\affiliation{
  \department{Institutes of Physical Science and Information Technology}
  \institution{Anhui University}
  \city{Hefei}
  \country{China}
}

\author{Siyu Song}
\affiliation{
  \department{School of Computer Science and Technology}
  \institution{East China Normal University}
  \city{Shanghai}
  \country{China}
}

\author{Ziwen Wang}
\affiliation{
  \department{School of Computer Science and Technology}
  \institution{Anhui University}
  \city{Hefei}
  \country{China}
}
\author{Shangshang Yang}
\affiliation{
  \department{School of Computer Science and Technology}
  \institution{Anhui University}
  \city{Hefei}
  \country{China}
}
\renewcommand{\shortauthors}{Yin Yi et al.}

\begin{abstract}
Knowledge tracing (KT) predicts whether a student will answer the next question correctly from their response history.
Newly launched educational platforms often provide question texts but lack response logs for training, motivating zero-shot cross-domain KT.
Existing transfer and cold-start approaches commonly leverage question semantics, but semantic discrepancies between platforms can limit the transferability of source-trained representations.
To address this challenge, we propose \textbf{TIZSKT}, a framework that separates behavioral sequence modeling from domain-local historical evidence construction, using text as an index rather than a directly transferable predictive representation.
The Mastery-Dynamics Encoder learns correctness dynamics without receiving question identifiers or text embeddings.
In parallel, Domain-Local Question Addressing constructs fixed question addresses using a frozen text encoder and an unsupervised per-domain projection.
The Text-Addressing Kernel uses these addresses to aggregate the current student's previous outcomes into a question-specific historical correctness summary.
Cross-Segment Continuation retains access to earlier responses without extending the behavioral encoder's input sequence.
Evidence-Volume Read-Out preserves the total kernel mass discarded by normalization and combines its log-compressed value with the behavioral state and historical correctness summary to predict the next response.
All trainable parameters are learned from source domains; target-domain inference uses public question texts and the current student's observed history without target-domain training.
Leave-one-domain-out experiments on three public datasets show that \textbf{TIZSKT} achieves the highest mean AUC and accuracy among the evaluated baselines on all target domains.
Ablation studies further demonstrate the contributions of text addressing and cross-segment continuation.
\end{abstract}


\maketitle

\section{Introduction}
\label{sec:introduction}

\begin{figure}[t]
    \centering
    \includegraphics[width=0.7\linewidth]{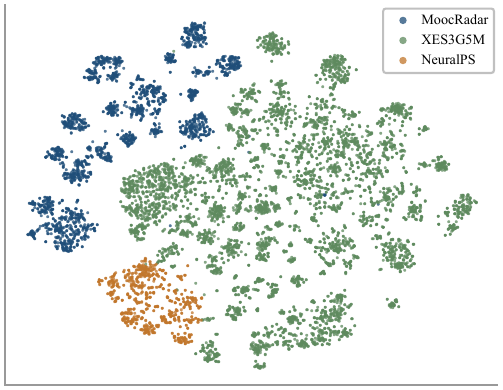}
    \caption{
        Visualization of frozen text representations for all
        questions across the three domains.
        Points are reduced for visualization alone: a
        256-dimensional PCA projection is fitted to the pooled
        question set, followed by t-distributed stochastic neighbor
        embedding.
    }
    \label{fig:domains}
    \vspace{-5mm}
\end{figure}

Knowledge tracing (KT) aims to predict whether a student will answer
the next question correctly based on their response history.
It supports personalized practice recommendation, learning
interventions, and instructional feedback on online education
platforms
\cite{corbett1994bkt,piech2015deep,abdelrahman2023knowledge}.
Mature platforms can train deep KT models on extensive interaction
records
\cite{ghosh2020context,liu2023simplekt,li2024stablekt}.
Newly launched platforms, however, often have only question texts
and lack sufficient response logs for training.
This motivates \emph{zero-shot cross-domain knowledge tracing}:
learning from established source domains to make predictions on a
new target domain without using target interactions for training
or model selection.
At inference time, the model can access public target question texts
and the current student's previously observed responses.
Thus, zero-shot refers to the absence of target-domain training,
rather than the absence of a student's history at inference.

Existing research addresses data scarcity through transfer learning
and question cold-start modeling, often using semantic information
to reduce dependence on domain-specific question identifiers.
Cross-domain KT methods exploit relationships between questions or
knowledge concepts to transfer information across courses and
platforms \cite{han2025transkt,xie2024domain}.
Question cold-start methods use language-model features to represent
new questions with few or no observed responses
\cite{guo2024mitigating,jung2025clst}.
Related zero-shot settings have also been explored in cognitive
diagnosis \cite{gao2024zero,liu2025lrcd} and sequential
recommendation
\cite{hou2022unisrec,li2023recformer,wang2024preprec},
although their prediction objectives and target-domain access
conditions differ from those considered here.
These studies motivate using question semantics to connect
otherwise disjoint question vocabularies, but do not by themselves
establish that a source-trained semantic transformation will remain
effective on a new KT domain.

A central challenge is the semantic discrepancy between the question
sets of different platforms.
Differences in question content and wording can produce distinct
representation distributions, even when all questions are encoded
by the same pretrained text encoder.
A predictor trained to interpret question representations on source
domains may therefore encounter a different semantic distribution
at deployment.
Without target-domain training, the model has limited opportunity
to adapt to this mismatch, and directly transferring learned
semantic transformations may degrade predictive performance.
Our empirical observations are consistent with this concern:
frozen question embeddings exhibit strong domain clustering
(\Cref{fig:domains}), while introducing text representations into
existing KT architectures produces gains that vary across
architectures and target domains (\Cref{tab:main}).
These observations motivate an alternative use of text that reduces
reliance on question embeddings as directly transferable predictive
inputs.

To address this challenge, we propose
\emph{text as an index, not a transfer carrier}.
The key is to distinguish learning response dynamics from
constructing question-specific historical evidence.
A behavioral encoder learns from correctness sequences, whose
binary input vocabulary is shared across platforms and does not
require question identifiers to be aligned.
Question text, in turn, provides addresses for relating the next
question to the current student's previous responses.
Since the query question and the historical questions belong to
the same domain, this retrieval does not require correspondence
between source and target questions or alignment of their
domain-local coordinate systems.
The model can therefore share the mechanisms for modeling response
dynamics and interpreting retrieved evidence while constructing
that evidence within each student's own domain.

We instantiate this design in TIZSKT
(Text-as-Index Zero-Shot Knowledge Tracing), which combines a
behavioral modeling branch, a history-addressing branch, and an
evidence-aware read-out.
Specifically, the Mastery-Dynamics Encoder (MDE) uses pointwise
aggregated attention \cite{zhai2024hstu} to encode the correctness
sequence of the current segment.
It receives neither question identifiers nor text embeddings,
separating the modeling of correctness dynamics from the
interpretation of question content.
In parallel, Domain-Local Question Addressing (DLQA) constructs
fixed question addresses from a frozen text encoder through a
principal-component projection fitted separately to each domain's
question texts, followed by normalization.
This address construction uses no response labels and remains
fixed during training and inference.

The Text-Addressing Kernel (TAK) uses these addresses to relate
the next question to questions previously answered by the same
student and aggregates their observed outcomes into a normalized
historical correctness summary.
The addressing formulation also accommodates a gated,
state-dependent contribution to relevance scores, allowing
historical addresses to be read against the evolving behavioral
state while the addresses themselves remain fixed.
The resulting summary complements the behavioral state with
question-specific evidence from past responses.
Both the query and the retrieved responses remain within the
current student's domain; no shared question identifiers or
explicit cross-domain alignment of question representations
is required.

Two further components address limitations of historical evidence
construction and read-out.
First, dividing response sequences into segments can exclude
earlier interactions that remain relevant to the next question.
Cross-Segment Continuation (CSC) extends the history available
to TAK by retaining a bounded window of responses from the same
student's preceding segments, together with the observed responses
in the current segment.
MDE continues to process only the current segment, so earlier
responses contribute to retrieval without extending the encoder's
input sequence.
Second, a normalized correctness summary does not retain its
total kernel mass: histories with different amounts of weighted
support can produce the same historical correctness rate.
Evidence-Volume Read-Out (EVRO) preserves this information through
the log-compressed sum of the unnormalized kernel weights.
Its prediction layer combines the behavioral state, historical
correctness summary, and evidence-volume feature to estimate
the probability of a correct response to the next question.
All trainable parameters are learned from source domains and
remain fixed during target-domain inference.

We evaluate TIZSKT on three public datasets---MoocRadar, NeuralPS,
and XES3G5M---using a leave-one-domain-out protocol.
Each experiment trains on two domains and tests on the remaining
domain, with model selection restricted to source-domain
validation data.
Across five random seeds, TIZSKT achieves the highest mean AUC
and accuracy among the evaluated baselines on all three target
domains (\Cref{sec:exp-main}).
Ablation studies further show that domain-local text addressing
and cross-segment continuation contribute to predictive performance.
Together, these results support combining source-trained
behavioral modeling with within-domain historical evidence for
zero-shot cross-domain KT.

\section{Related Work}
\label{sec:related-work}

\subsection{Knowledge Tracing}
\label{sec:rw-kt}

Knowledge tracing estimates a student's evolving knowledge state
from past interactions to predict future responses
\cite{abdelrahman2023knowledge}. Deep models learn this temporal
dependence through sequence modeling and attention mechanisms
\cite{piech2015deep,ghosh2020context,liu2023simplekt,li2024stablekt}.
Alongside advances in student modeling, question representations have
been enriched through pre-training on question--concept bipartite
graphs \cite{liu2020improving} and automated concept annotation
\cite{ozyurt2025automated}. A central challenge is how to exploit
question content when interaction histories provide insufficient
evidence for learning question representations.

Recent work explores LLMs both as response predictors and as a source
of information for cold-start questions. Studies of direct prediction
investigate deduction from perturbed evidence
\cite{sonkar2023deduction}, instruction-based alignment
\cite{wang2025llmkt}, and student behavior modeling
\cite{norris2025next,moon2025using}. Reported comparisons show that
specialized models can achieve higher accuracy at lower computational
cost \cite{bhattacharyya2026faster,cho2024systematic}.
To address question cold start, existing methods introduce
attribute-aware LLM features \cite{guo2024mitigating}, kernel bias
with cone attention \cite{bai2025cskt}, generative models aligned
for knowledge tracing \cite{jung2025clst}, and attributes-enhanced
representation learning \cite{yu2026attributes,lee2025enhancing}.
These approaches use question semantics to support prediction when
question-specific response evidence is unavailable.

\subsection{Cross-Domain Transfer}
\label{sec:rw-zsrec}

Cross-domain knowledge tracing studies how models trained on one
course or platform can support prediction in another
\cite{han2025transkt,xie2024domain,lin2026mckt,duan2026ragkt}.
Beyond the absence of responses for individual questions, this
setting raises the question of whether representations learned in
one domain remain useful in another. Related work in cognitive
diagnosis investigates transfer through an early batch of target
students \cite{gao2024zero}, language-cognitive mappings
\cite{liu2025lrcd}, and zero-shot LLM diagnosticians
\cite{ma2025large}. Among these methods, LRCD \cite{liu2025lrcd} performs transfer without target-domain training and
represents entities from different domains in a shared text-derived
space. Its textual profiles are constructed from target-domain
response logs at inference, and its mapping is applied once per
entity to estimate a static student mastery vector. Knowledge tracing
requires predictions conditioned on an evolving interaction history,
which introduces a temporal dimension beyond this static formulation.

Sequential recommendation also models evolving interaction histories,
using text-based item representations
\cite{li2023recformer} and semantic alignment
\cite{li2025llmrecg}. Transfer in this setting can involve additional
target-domain information. UniSRec \cite{hou2022unisrec} fine-tunes adapters on target
sequences, while PrepRec uses target-domain popularity statistics
\cite{wang2024preprec}. Together with cross-domain student modeling,
these approaches motivate studying how question semantics can support
sequential prediction without target-domain training or population
statistics. We investigate whether text can guide retrieval within
a student's own history, allowing transfer to rely on behavioral
patterns rather than a shared semantic representation across domains.
\section{Problem Definition}
\label{sec:problem-definition}

Knowledge tracing (KT) estimates whether a student will answer a
question correctly from their interaction history \cite{corbett1994bkt}:
a student $u$ on a platform leaves a sequence
$X_u = ((q_1, a_1), \dots, (q_{n_u}, a_{n_u}))$, where $q_i$ identifies
the question and $a_i \in \{0,1\}$ records the outcome. A domain $d$ is
one dataset: a pool of questions $\mathcal{Q}^{(d)}$ carrying the text
released with its dataset, and the interaction sequences of its own
students.

Given the history $X_u$ and a next question $q$, the model outputs a
probability
\begin{equation}
\hat{y} = f_\theta(X_u, q) \in [0,1], \qquad
\hat{y} \approx \Pr(a = 1 \mid X_u, q),
\label{eq:kt}
\end{equation}
scored by AUC on a held-out test split. We study KT under domain shift,
evaluated leave-one-domain-out over domains with disjoint student
populations: for each target $t$, the remaining domains form the training
dataset and the model is scored on $t$ alone---the cold-start situation of
a platform that has published its question texts but has collected no
interaction.

\label{sec:protocol}
Zero-shot KT removes the target domain's interactions from training
entirely: parameters are fitted on source-domain interactions and texts,
and no target-domain interaction, nor any statistic computed from one,
enters training or model selection. The target domain enters only
through its public question texts, each encoded once by a frozen
pre-trained encoder without gradient, so that a domain's texts index
only that domain; the target student's own history is read as inference
input, exactly as the task definition requires. Everything the model
consumes from text is computed inside the domain that consumes it---the
constraint \Cref{sec:overview} turns into a design principle.

\section{Methods}
\label{sec:methods}

\begin{figure*}[t]
\centering
\includegraphics[width=0.8\textwidth]{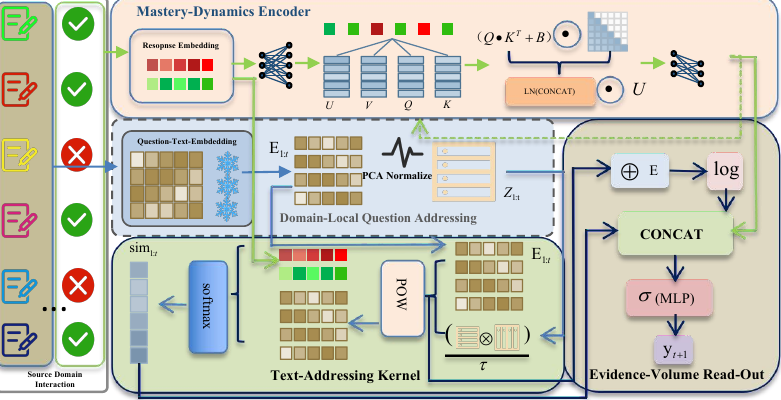}
\caption{TIZSKT consists of the Mastery-Dynamics Encoder, Domain-Local Question Addressing , a Text Addressing Kernel , Cross-Segment Continuation
and Evidence-Volume Read-out. }
\label{fig:architecture}
\vspace{-3mm}
\end{figure*}

\textbf{TIZSKT} combines behavioral sequence modeling with domain-local
text addressing for zero-shot cross-domain knowledge tracing.
As shown in \Cref{fig:architecture}, the model has two complementary
branches. Domain-Local Question Addressing (\textbf{DLQA}) assigns every
question a fixed address inside its own domain, obtained from text alone
and without gradient. The Mastery-Dynamics Encoder (\textbf{MDE}) learns
from correctness sequences using an input vocabulary shared across
domains. The history-addressing branch uses those addresses to retrieve
relevant responses from the current student's history.
Within this branch, the Text-Addressing Kernel (\textbf{TAK}) summarizes
historical correctness, Cross-Segment Continuation (\textbf{CSC}) expands
the available history, and Evidence-Volume Read-Out (\textbf{EVRO}) preserves
the kernel mass discarded by normalization. A prediction layer
combines the behavioral state and the retrieved evidence. Specific details are presented in the subsequent chapters.
\label{sec:overview}

\subsection{Domain-Local Question Addressing}
\label{sec:dlqa}
Question-specific prediction therefore requires a way to
relate the upcoming question to the student's experience, and we build
that relation inside each domain rather than depend on a source-trained
transformation of question semantics.

The only target-side information the protocol admits is question text,
in the language space of a frozen encoder, and zero-shot cognitive
diagnosis answers the same requirement with a learned
language--cognitive mapper \cite{liu2025lrcd}. Such a mapper is fitted on
source domains and must then be transported across the boundary this
protocol refuses to cross. The conversion here is therefore
closed-form. Every question of a domain is encoded once, without
gradient, by the frozen Qwen3-Embedding-8B encoder into
$\mathbf{e}_q \in \mathbb{R}^{4096}$, and projected by a
principal-component basis $\mathbf{B}_d \in \mathbb{R}^{4096 \times d_z}$
fitted on that domain's own question texts:
\begin{equation}
\mathbf{z}_q
=
\frac{
\mathbf{B}_d^{\top}(\mathbf{e}_q-\boldsymbol{\mu}_d)
}{
\left\|
\mathbf{B}_d^{\top}(\mathbf{e}_q-\boldsymbol{\mu}_d)
\right\|_2
},
\label{eq:pca}
\end{equation}
where $\boldsymbol{\mu}_d$ is the mean frozen embedding over the
domain's pool. No whitening is applied, so the relative scale of the
components survives. Because the vectors are $\ell_2$-normalized, an
address inner product is a cosine similarity, and because the basis is
estimated from question text alone---no gradient, no response label---the
construction runs identically on a target domain as soon as its texts
become public and stays fixed through training and through target-domain
inference alike. The projection is a change of space, from the language
space of the frozen encoder to the one in which this model compares
items, and it is the only conversion the model performs.

Estimating that space per domain is also what makes comparison between
domains unnecessary: a query and the history it is scored against are
always expanded in the same basis, so retrieval needs no alignment
between coordinate systems and every similarity is computed and consumed
inside the single domain that produced it. What this section hands to the
rest of the model is consequently not a transported representation but an
index---a table of unit vectors that the following sections read by
cosine.

\Cref{eq:pca} is a function of text alone, which suffices where the
cognitive object is one static mastery vector. Knowledge tracing
estimates a trajectory instead, so the conversion was also given a
learned half that re-reads each address against the evolving state,
behind a scalar gate initialized to zero:
\begin{equation}
s_j
=
\frac{\cos(\mathbf{z}_{q^\star}, \mathbf{z}_{q_j})}{\tau}
+ \alpha \cdot
\frac{\langle \mathbf{v}_t, \mathbf{u}_{q_j} \rangle}{\tau}
+ \ln\lambda \cdot (t-j),
\qquad
\label{eq:dlqa-soft}
\end{equation}
where $\mathbf{u}_q$ and $\mathbf{v}_t$ are $\ell_2$-normalized images of
the address and of the state under two shared two-layer ReLU networks.
Since $\alpha$ is one scalar starting at zero, the gated model coincides
with \Cref{eq:pca} at initialization and differs from it in exactly one
degree of freedom, so the comparison isolates that channel and nothing
elses.

\subsection{Mastery-Dynamics Encoder}
\label{sec:ba}

Addresses fix how questions are compared; what remains is to summarize
what the student has done. To keep this question-free, MDE receives
correctness labels and nothing else. Their shared two-symbol alphabet
supports learning response patterns across platforms, while the
question-specific information is supplied by retrieval, so the modeling
of correctness dynamics is separated from the interpretation of
question content.

Each response $j\in\mathcal{C}_t$ is mapped to the encoder space:
\begin{equation}
\mathbf{x}^{(0)}_j
=
W_r\,\mathrm{Emb}(a_j),
\qquad
\mathrm{Emb}(a_j)\in\mathbb{R}^{d_{\mathrm{model}}},
\label{eq:resp-emb}
\end{equation}
where $\mathrm{Emb}$ is a learned correctness embedding and
$W_r\in\mathbb{R}^{d_{\mathrm{model}}\times d_{\mathrm{model}}}$
is a trainable projection. The embeddings form the rows of
$X^{(0)}$ in chronological order.

MDE contextualizes individual responses using $L$ blocks with
$H$ attention heads. Relative-position biases encode the temporal
relations between outcomes, while pointwise aggregated attention
\cite{zhai2024hstu} combines their features through gated updates:

\begin{align}
U^{(\ell)},V^{(\ell)},Q^{(\ell)},K^{(\ell)}
&=
\mathrm{Split}\!\left(
\phi\!\left(\mathrm{RMSNorm}(X^{(\ell)})W_1^{(\ell)}\right)
\right),
\label{eq:enc-qkvu}\\
A^{(\ell,r)}
&=
M\odot\phi\!\left(
Q^{(\ell,r)}(K^{(\ell,r)})^\top+b^{(\ell,r)}
\right),
\label{eq:enc-attn}\\
X^{(\ell+1)}
&=
X^{(\ell)}
+
\left[\mathrm{LN}(C^{(\ell)})\odot U^{(\ell)}\right]
W_2^{(\ell)},
\label{eq:enc-out}
\end{align}
where $\ell$ indexes blocks, $r$ indexes heads, and $\phi$ is
the sigmoid linear unit (SiLU). RMSNorm and LN denote root mean
square normalization and layer normalization, respectively.
The projections $V^{(\ell)},Q^{(\ell)},K^{(\ell)}$ are split
into heads, and $C^{(\ell)}$ concatenates
$A^{(\ell,r)}V^{(\ell,r)}$ over $r$.
The gate $U^{(\ell)}$ matches the dimensions of $C^{(\ell)}$,
$W_2^{(\ell)}$ maps the gated features back to
$d_{\mathrm{model}}$, and $\odot$ denotes elementwise multiplication.
Because $\phi$ acts elementwise rather than across a row, no
renormalization is introduced and an aggregate scales with the amount of
contributing evidence instead of being forced into a convex combination;
the contrast matters at \Cref{sec:evr}.

Each entry of $b^{(\ell,r)}$ is indexed by a relative-distance bucket,
a zero-initialized table of $B{=}16$ buckets carrying one scalar per
head, addressed by a fixed log-spaced function of the distance
$\delta = i-j$ alone:
\begin{equation}
\beta(\delta) = \begin{cases}
\delta, & \delta < B/2,\\[2pt]
\left\lfloor \dfrac{B}{2} +
\dfrac{\log\!\big(\delta/(B/2)\big)}{\log\!\big(512/(B/2)\big)}\,
\dfrac{B}{2} \right\rfloor, & \delta \ge B/2.
\end{cases}
\label{eq:buckets}
\end{equation}
Distances inside $B/2$ receive a bucket each, so exact order is retained
at short range, while longer ones are compressed logarithmically. Relative
rather than absolute position keeps the encoder invariant to where a
segment begins, the invariance \Cref{sec:continuation} exploits, and
zero initialization starts the blocks with no positional bias at all.

For a query at position $i$, the mask $M$ retains only valid keys
$j\leq i$ within the current segment. Padding positions are excluded from
attention and the loss. The final representation of response $a_t$,
\begin{equation}
\mathbf{h}_t
=
\mathrm{RMSNorm}\big(\mathbf{x}^{(L)}_t\big)
\in\mathbb{R}^{d_{\mathrm{model}}},
\label{eq:final-state}
\end{equation}
summarizes the correctness dynamics of the current segment.

\subsection{Text-Addressing Kernel}
\label{sec:ta}

Addresses describe relations between questions, whereas prediction
requires evidence about the student's performance. TAK connects
them by weighting historical correctness according to similarity
to the upcoming question. This converts domain-local semantic
relations into a response statistic with the same definition
across platforms.

For each history index $j\in\mathcal{J}_t$, the kernel computes:
\begin{equation}
s_{t,j}
=
\frac{
\mathbf{z}_{q^\star}^{\top}\mathbf{z}_{q_j}
}{\tau},
\qquad j\in\mathcal{J}_t,
\label{eq:kernels}
\end{equation}
where $\tau>0$ is a fixed temperature controlling weight
concentration. The kernel applies no explicit temporal decay;
recency constrains the history through CSC.
The normalized scores yield the historical correctness summary:
\begin{equation}
\mathrm{sim}_t
=
\begin{cases}
\displaystyle
\frac{
\sum_{j\in\mathcal{J}_t}e^{s_{t,j}}a_j
}{
\sum_{j\in\mathcal{J}_t}e^{s_{t,j}}
},
& \mathcal{J}_t\neq\varnothing,\\[8pt]
\frac{1}{2},
& \mathcal{J}_t=\varnothing.
\end{cases}
\label{eq:normalized}
\end{equation}
Read as an estimator, this is a Nadaraya--Watson regression of the
correctness label against position in address space, with the kernel
supplying the local weighting and $\tau$ the inverse bandwidth. The query
and the keys come from the same domain by construction, so nothing in
\Cref{eq:kernels} has to survive the domain boundary. This
parameter-free estimate emphasizes outcomes on questions
similar to $q^\star$, with a fixed uninformative prior for empty
histories. It complements $\mathbf{h}_t$ by conditioning historical
performance on the next question.

\subsection{Cross-Segment Continuation}
\label{sec:continuation}

Similarity weighting can recover relevant evidence only from
the history available to it. Computational segment boundaries
need not coincide with changes in student knowledge, yet
current-segment retrieval would discard all earlier responses.
CSC therefore separates the retrieval horizon from MDE's context.

Let $\mathcal{H}_t$ contain valid observed interaction indices
up to time $t$ for the current student and domain within the
current data split. CSC selects the retrieval history as:
\begin{equation}
\mathcal{J}_t
=
\operatorname{Recent}_W(\mathcal{H}_t),
\label{eq:history-window}
\end{equation}
where $\operatorname{Recent}_W$ retains the interactions preceding the
current segment up to a window of $W$ items, together with everything the
student has already answered inside it. The bound therefore applies to
the student's preceding segments only, never across a train--test
boundary, and $|\mathcal{J}_t|$ may exceed $W$ by the length of the
current segment prefix.

The construction is identical during training and evaluation.
MDE continues to use $\mathcal{C}_t$, while TAK and EVRO operate
on $\mathcal{J}_t$. Earlier responses thus contribute
question-specific evidence without extending the trainable
encoder's input sequence.

\subsection{Evidence-Volume Read-Out}
\label{sec:evr}

CSC expands the accessible history, but TAK's normalized average
does not retain its total kernel mass. Different histories can
therefore produce the same correctness rate despite having
different amounts of weighted support. EVRO preserves the
discarded normalization factor through:
\begin{equation}
\mathrm{EV}^{\mathrm{sim}}_t
=
\log\left(
1+\sum_{j\in\mathcal{J}_t}e^{s_{t,j}}
\right).
\label{eq:ev}
\end{equation}
The feature depends on both history size and query--history
similarities. Logarithmic compression reduces its dynamic range,
and an empty history yields zero. TAK and EVRO reuse the same
masked kernel scores.

The read-out combines the current correctness dynamics,
question-specific historical performance, and its kernel support:
\begin{equation}
\hat{y}_{t+1}
=
\sigma\left(
\mathrm{MLP}\left(
\left[
\mathbf{h}_t;\,
\mathrm{sim}_t;\,
\mathrm{EV}^{\mathrm{sim}}_t
\right]
\right)
\right),
\label{eq:readout}
\end{equation}
where $[\,\cdot;\cdot\,]$ denotes concatenation and $\sigma$ is
the sigmoid function. This interface allows source supervision
to learn how behavioral state and retrieved evidence interact,
while the evidence itself is constructed within the student's
domain. The first-layer weights associated with
$\mathrm{EV}^{\mathrm{sim}}_t$ are initialized to zero and
subsequently learned. With identical remaining parameters,
initial predictions match the read-out without this feature.

\subsection{Training Objective}
\label{sec:training}

Training minimizes binary cross-entropy over source-domain
prediction steps. Each batch is encoded twice; the passes differ only in
their dropout draws, so they are two views of one input. Let
$\mathbf{z}^{(k)}$ be the logits of pass $k \in \{1,2\}$, let
$\mathrm{bce}^{(k)}_{u,t}$ be its binary cross-entropy at step $(u,t)$,
and let $\mathcal{M}$ be the valid prediction steps of the batch, with
$\mathcal{M}_d$ its restriction to the students of source domain $d$.
Two auxiliary quantities accompany the mean loss. First,
\begin{equation}
\bar{\ell}^{(k)}_{d}
=
\mathbb{E}_{(u,t)\in\mathcal{M}_d}
\big[\mathrm{bce}^{(k)}_{u,t}\big],
\qquad
\mathcal{V}^{(k)}
=
\mathrm{Var}_{d\in\mathcal{D}}
\big(\bar{\ell}^{(k)}_{d}\big),
\label{eq:risk}
\end{equation}
the mean loss within each source domain of the batch and the dispersion
of those means across the source domains $\mathcal{D}$ present. Second,
\begin{equation}
\mathcal{L}_{\mathrm{cons}}
=
\mathbb{E}_{(u,t)\in\mathcal{M}}
\Big[
\big(z^{(1)}_{u,t}-z^{(2)}_{u,t}\big)^2
\Big],
\label{eq:cons}
\end{equation}
the disagreement between the two views of the same batch. The objective
is
\begin{equation}
\mathcal{L}
=
\frac{1}{2}\sum_{k=1}^{2}
\Big[
\mathbb{E}_{(u,t)\in\mathcal{M}}
\big[\mathrm{bce}^{(k)}_{u,t}\big]
+
\lambda_{\mathrm{risk}}\,\mathcal{V}^{(k)}
\Big]
+
\lambda_{\mathrm{cons}}\,\mathcal{L}_{\mathrm{cons}},
\label{eq:kt-loss}
\end{equation}

\section{Experiments}
\label{sec:experiments}

\subsection{Experimental Setting}
\label{sec:exp-setup}

\paragraph{Dataset.}
We conduct our experiments on three public datasets: MoocRadar
\cite{yu2023moocradar}, a course-scale dataset with the most decidedly
positive answer rate; NeuralPS, our extraction of the TASK34
answer-prediction subset of the NeurIPS 2020 Education Challenge
\cite{wang2020instructions}, the smallest pool and the only one close to
an unbiased answer rate; and XES3G5M \cite{liu2023xes3g5m}, a mathematics
benchmark from a real-world platform for third-grade students and the
largest by every count. \Cref{tab:datasets} reports the size of each
dataset---the number of questions, students, and interaction
records---together with its answer base rate and the held-out target
test split used in the leave-one-domain-out assignment. Each dataset is
partitioned \emph{by student} into training, validation, and test parts
in the ratio $7{:}1{:}2$, so no student's responses straddle a split.

\begin{table}[t]
\caption{Datasets and held-out target test splits under the
leave-one-domain-out assignment.}
\label{tab:datasets}
\small
\setlength{\tabcolsep}{4pt}
\begin{tabular}{@{}lrrrrr@{}}
\toprule
Dataset & Questions & Students & Interactions & Correct & Test int. \\
\midrule
MoocRadar & 2{,}513 & 14{,}224 & {\textasciitilde}0.9M & 81.7\% & 178{,}600 \\
NeuralPS & 948 & 4{,}918 & 1.38M & 53.0\% & 284{,}701 \\
XES3G5M & 7{,}651 & 18{,}066 & 5.55M & 79.5\% & 1{,}105{,}018 \\

\bottomrule
\end{tabular}
\vspace{-3mm}
\end{table}
\vspace{-3mm}

\paragraph{Training and evaluation.}
For each leave-one-domain-out transfer, we train on the two source
domains' training sequences and use their validation sets for model
selection; no target-domain statistic enters selection. We select the
model weights at the best source-validation checkpoint and evaluate
those weights on the target domain's held-out test split, where the
target student's own history serves as inference input exactly as
\Cref{eq:kt} requires. We report AUC and
ACC on the target test. Every
reported number is the mean and sample standard deviation over five
random seeds.

Four KT architectures of contrasting design run under two zero-shot
adaptations, five seeds each: AKT \cite{ghosh2020context} with
Rasch-style factorized representations and monotone decay; RobustKT
\cite{guo2025robustkt} with a cognitive decoupling analyzer that
separates stable cognitive patterns from random factors by smoothing
and subtraction; and stableKT \cite{li2024stablekt}, which replaces
positional encoding with linear relative biases. \emph{Random IDs} keeps
learnable question identifiers, so target-domain identifiers stay at
random initialization---the identity-noise control. \emph{Text-Rasch}
replaces the question embedding with the frozen text table and a
trainable projection and conditions the Rasch difficulty factors on
text through a small MLP, the strongest semantic adaptation we could
construct; DKT has no such variant, its full-question output head being
undefined over unseen identifiers. Five cross-domain and graph-based
baselines run under the same access rules and frozen addresses: TransKT
\cite{han2025transkt}, which embeds questions and concept names and
builds its concept graph from LLM-judged relations; SINKT
\cite{fu2024sinkt}; the two EKT encoders \cite{liu2021ekt}; and DyGKT
\cite{cheng2024dygkt}, whose dynamic graph binds to one dataset's own
nodes.

The \emph{Embedding input} column of \Cref{tab:main} names what each row
feeds its question representation: \texttt{random} is the Random-IDs
control, \texttt{Qwen3-8B} the Text-Rasch adaptation, \texttt{EKT-native}
EKT's own word-level encoder, and \texttt{IDs} concept identifiers with
no question text at all. Every row that consumes text consumes the
identical frozen Qwen3-Embedding-8B signal of
\Cref{sec:problem-definition}, so the comparison isolates what each
architecture asks of the text rather than which encoder it uses. All
trained models are implemented in PyTorch and trained on
NVIDIA Tesla T4 GPUs.

All trained models share one configuration: encoder dimension $256$, two encoder blocks over eight heads, 16 relative-bias buckets, kernel temperature $\tau{=}0.25$, decay
$\lambda{=}1$ (disabled), and a continuation window of $200$ items. Each
run trains for $30$ epochs at batch size $64$ in single precision under
\Cref{eq:kt-loss} . with $\lambda_{\mathrm{risk}}{=}0.1$ and
$\lambda_{\mathrm{cons}}{=}0.02$. Setting
$\lambda_{\mathrm{risk}}{=}\lambda_{\mathrm{cons}}{=}0$ recovers plain
cross-entropy; Both auxiliary terms read source-domain labels only, and
neither is active at target-domain inference, where a single
deterministic pass is used, so neither touches the target side of the
protocol.

\subsection{Overall Performance}
\label{sec:exp-main}

\begin{table*}[t]
\caption{Target-domain test AUC and ACC, mean $\pm$
sample standard deviation over five seeds. Bold: best per column.}
\label{tab:main}
\small
\newcommand{\sd}[1]{{\small$\pm$#1}}
\setlength{\tabcolsep}{4pt}
\renewcommand{\arraystretch}{0.80}
\begin{tabular}{@{}llcccccc@{}}
\toprule
& & \multicolumn{2}{c}{MoocRadar} & \multicolumn{2}{c}{NeuralPS} & \multicolumn{2}{c}{XES3G5M} \\
\cmidrule(lr){3-4}\cmidrule(lr){5-6}\cmidrule(lr){7-8}
Method & Embedding & AUC & ACC & AUC & ACC & AUC & ACC \\
\midrule
AKT \cite{ghosh2020context} & random & 0.7814\sd{0.0005} & 0.8245\sd{0.0006} & 0.7207\sd{0.0007} & 0.6605\sd{0.0023} & 0.6943\sd{0.0078} & 0.7885\sd{0.0077} \\
AKT \cite{ghosh2020context} & Qwen3-8B & 0.7804\sd{0.0007} & 0.8171\sd{0.0021} & 0.7200\sd{0.0005} & 0.6351\sd{0.0323} & 0.6918\sd{0.0023} & 0.7804\sd{0.0151} \\
RobustKT \cite{guo2025robustkt} & random & 0.7927\sd{0.0066} & 0.8259\sd{0.0011} & 0.7159\sd{0.0017} & 0.6439\sd{0.0074} & 0.6829\sd{0.0031} & 0.7843\sd{0.0026} \\
RobustKT \cite{guo2025robustkt} & Qwen3-8B & 0.8020\sd{0.0009} & 0.8273\sd{0.0018} & 0.7189\sd{0.0014} & 0.6535\sd{0.0054} & 0.6845\sd{0.0027} & 0.7928\sd{0.0081} \\
stableKT \cite{li2024stablekt} & random & 0.7823\sd{0.0004} & 0.8235\sd{0.0019} & 0.7203\sd{0.0004} & 0.6620\sd{0.0016} & 0.7001\sd{0.0012} & 0.7913\sd{0.0035} \\
stableKT \cite{li2024stablekt} & Qwen3-8B & 0.7820\sd{0.0006} & 0.8213\sd{0.0049} & 0.7211\sd{0.0003} & 0.6450\sd{0.0175} & 0.7013\sd{0.0009} & 0.7817\sd{0.0151} \\
\midrule
TransKT \cite{han2025transkt} & Qwen3-8B & 0.7951\sd{0.0011} & 0.8251\sd{0.0015} & 0.7115\sd{0.0023} & 0.6430\sd{0.0057} & 0.7031\sd{0.0006} & 0.7757\sd{0.0035} \\
EKT-Attn \cite{liu2021ekt} & EKT-native & 0.7324\sd{0.0051} & 0.8055\sd{0.0001} & 0.6708\sd{0.0076} & 0.5553\sd{0.0005} & 0.6516\sd{0.0019} & 0.7870\sd{0.0001} \\
EKT-EKTA \cite{liu2021ekt} & EKT-native & 0.7547\sd{0.0039} & 0.8182\sd{0.0008} & 0.6919\sd{0.0053} & 0.6362\sd{0.0045} & 0.5937\sd{0.0559} & 0.6667\sd{0.0696} \\
SINKT \cite{fu2024sinkt} & Qwen3-8B & 0.7171\sd{0.0067} & 0.7912\sd{0.0129} & 0.7032\sd{0.0069} & 0.6463\sd{0.0039} & 0.6780\sd{0.0065} & 0.7556\sd{0.0188} \\
DyGKT \cite{cheng2024dygkt} & random & 0.4816\sd{0.0037} & 0.8092\sd{0.0142} & 0.4913\sd{0.0075} & 0.5504\sd{0.0000} & 0.4983\sd{0.0052} & 0.3027\sd{0.0720} \\
DyGKT \cite{cheng2024dygkt} & Qwen3-8B & 0.4981\sd{0.0276} & 0.6843\sd{0.2671} & 0.5035\sd{0.0047} & 0.5504\sd{0.0000} & 0.5045\sd{0.0054} & 0.3169\sd{0.1176} \\
\midrule
TIZSKT & Qwen3-8B & \textbf{0.8069}\sd{0.0004} & \textbf{0.8277}\sd{0.0004} & \textbf{0.7312}\sd{0.0010} & \textbf{0.6647}\sd{0.0017} & \textbf{0.7171}\sd{0.0007} & \textbf{0.8015}\sd{0.0005} \\
\bottomrule
\end{tabular}
\end{table*}

\Cref{tab:main} reports target-domain test AUC and accuracy (ACC). Three
observations follow. First, semantic carriers lose on both sides of the comparison. Within
all four KT architectures, replacing random question identifiers with
frozen question semantics does not lift the architecture: the
text-initialized variant of AKT sits at or below its random-initialized
counterpart on every domain, and where the swap helps on RobustKT and
stableKT the margin sits inside the five-seed spread. The cross-domain
baselines lose from the other direction. TransKT, the most elaborate
semantic carrier in the field, leads on one domain but is dominated on
the other two by architectures that merely index text. DyGKT, whose
graph binds to one dataset's own nodes, collapses to the chance level;
even handed the same Qwen3-8B table it stays within half an AUC point of
chance on every domain. EKT and SINKT, the two remaining semantic
consumers, trail the strongest baseline on every domain.

Second, these losses follow from a property of the representation
rather than of any architecture. The frozen embeddings of the three
datasets are almost perfectly separable by domain (\Cref{fig:domains}),
so a text vector is a usable address inside its domain and a poor
carrier across it, and every configuration in the table differs only in
what it asks that vector to do. What degrades is therefore not a
component but a route---the one that asks a question representation to
survive the transfer. Architectures that index text are unaffected,
because an index is read only within a domain; carriers are affected,
and the steps of that route fail independently of one another. DKT and DyGKT collapse for structural reasons---an
undefined output head over unseen identifiers, a graph bound to one
dataset's nodes---which says nothing about the quality of the
architecture underneath.

Finally, we lead every domain in both metrics, and the reason is
structural. Our trainable parameters never consume text, so no fitted
weight is distorted by the domain separation the frozen embeddings
carry, and our kernels consume text only as an index over the student's
own history inside the target domain. The division of labor is explicit
inside the model: the mastery-dynamics encoder carries the transfer
through correctness dynamics that speak the same alphabet in every
domain, the text-addressing kernel supplies the question-specific
evidence, cross-segment continuation extends that evidence to earlier
sessions, and the evidence-volume read-out tells the prediction layer
how much evidence backs the summary. \Cref{sec:exp-ablation} measures
each of these contributions separately. The losses also fall unevenly,
so the strongest zero-shot baseline differs on every domain: a RobustKT
configuration leads on MoocRadar, stableKT on NeuralPS, and TransKT on
XES3G5M. A method that won by transporting semantics would have to beat
that moving reference consistently; none does, and ours does.

\subsection{Sensitivity Analysis}
\label{sec:exp-sensitivity}

\begin{figure*}[t]
\centering
\includegraphics[width=0.8\textwidth]{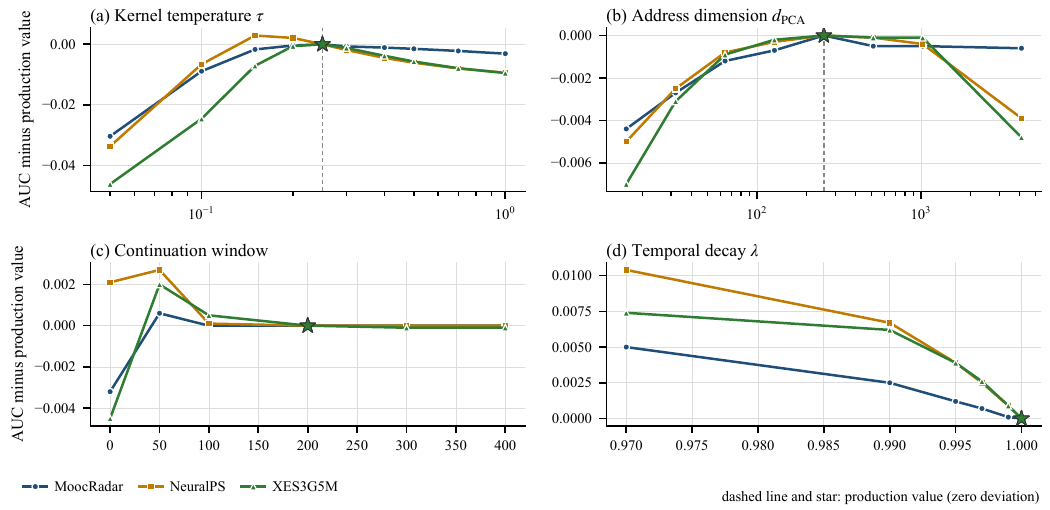}
\caption{Sensitivity of the zero-training addressing predictor, plotted as
the deviation of target-domain AUC from each domain's production value,
one line per domain; exact recomputation, no seed variance. Dashed line
and star: production value.}
\label{fig:sensitivity}
\vspace{-3mm}
\end{figure*}

\Cref{fig:sensitivity} sweeps the four protocol quantities on the
zero-training addressing predictor, where every point is an exact
recomputation rather than a training run. Two observations follow.
First, the surfaces are flat where it matters: the production
temperature sits within three thousandths of an AUC point of every
domain's optimum, the production continuation window sits on a plateau
that spans a four-fold range of window sizes on every domain, and the
address dimension plateaus across a sixteen-fold range of compression
settings on two of the three domains and a thirty-two-fold range on the
third, with the uncompressed setting falling below that plateau on two
of three---compression denoises rather than discards. This flatness is
why all four values are fixed rather than tuned. Second, where the
offline predictor is not flat it disagrees with the trained model: the
offline kernel prefers a shorter continuation window than the adopted
one on all three domains and light temporal decay on all three , and the one trained continuation probe moves the
opposite way, preferring a longer window than the adopted one by
margins inside seed noise . The trained model is the only arbiter, so the decay
ships disabled and the family exposes no temporal hyperparameter, while
the continuation window stays where both proxies agree.

\subsection{Ablation Studies}
\label{sec:exp-ablation}

We abbreviate the modules of \Cref{sec:methods} as DLQA (domain-local
question addressing), MDE (mastery-dynamics encoder), TAK
(text-addressing kernel), CSC (cross-segment continuation), and EVRO
(evidence-volume read-out). Each removal switches one module off and
re-trains from scratch under the same training protocol: \emph{w/o TAK}
drops the summary $\mathrm{sim}_t$, so the prediction layer reads the
behavioral state alone; \emph{w/o CSC} drops cross-segment continuation,
so the kernel reads the current segment only; \emph{w/o DLQA} removes
the similarity term from the score matrix, so $\mathrm{sim}_t$
degenerates to the mean historical correctness.

\begin{figure*}[t]
\centering
\includegraphics[width=0.70\textwidth]{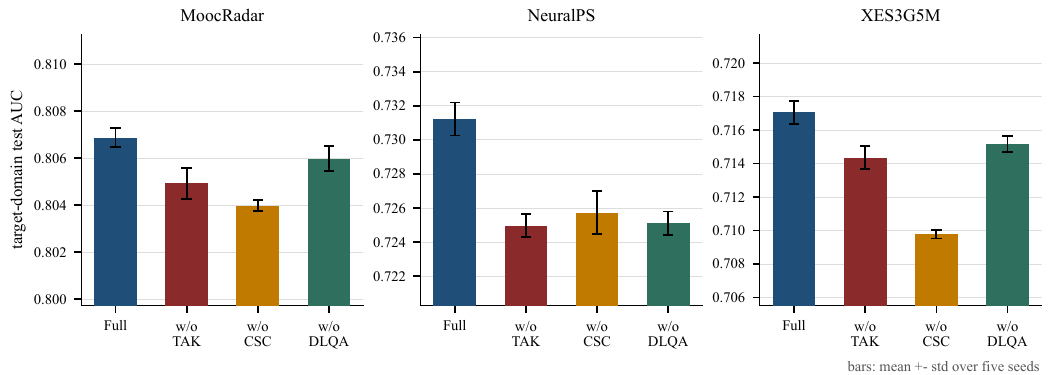}
\caption{Target-domain test AUC of the full model and of three removal
variants.}
\label{fig:ablation}
\vspace{-3mm}
\end{figure*}

\Cref{fig:ablation} turns each module off one at a time, and every
module of the adopted architecture earns its place: each removal lowers
target-domain test AUC on all three datasets, significantly under the
paired five-seed test. The removals nonetheless pay in different
places. Removing the addresses hurts NeuralPS most: the dataset sits
closest to an unbiased base rate, so the similarity-weighted summary is
the only question-specific signal the model has, and without the
addresses it degenerates to the running average; the addresses and the
kernel are the two ends of one pathway, and NeuralPS charges them
almost identically. Removing continuation hurts the other two datasets
most, where the current segment alone is thin evidence and the earlier
segments hold the student's history; NeuralPS, whose sequences stand by
themselves, misses it least. Removing the kernel sits between the other
two on all three datasets, so no removal is uniformly deepest: the
addressing pathway dominates where the base rate is uninformative, the
continuation window where segments are short.

The recency summary is the one component the adopted design omits:
re-adding it moves AUC by less than two thousandths on every domain,
and once trained the behavioral encoder subsumes the running-correctness
prior it supplies; the pointwise encoder with the evidence-volume
read-out likewise improves on the softmax variant on every domain. The
complementary half of the attribution falsifies the semantic-carrier
route inside the trainable pathway---learned transforms anti-transfer,
adversarial removal leaves the separation intact, recodings cost more
within-domain than they gain across domains, graph propagation acts as
an identity map, and learned gates close themselves---so semantic
information enters this problem productively only as an index, the conclusion \Cref{sec:dlqa} arrives at from
the opposite direction.
\vspace{-3mm}
\subsection{Case Study}
\label{sec:exp-cases}

\begin{figure}[t]
\centering
\includegraphics[width=0.8\linewidth]{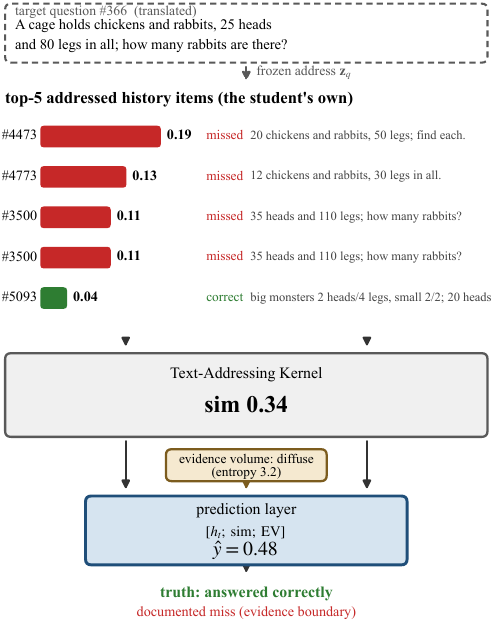}
\caption{Decision walkthrough, XES3G5M student 8099.}
\label{fig:case-study}
\vspace{-3mm}
\end{figure}

The case-study protocol samples, under pre-registered quotas, students
whose text-addressed evidence diverges from their overall record, and
inspects the model's decision against the ground truth.

\Cref{fig:case-study} walks one such student end to end. The target
asks how many rabbits sit behind 25 heads and 80 legs---arithmetically
the problem this student has already failed. The addressing kernel
keeps five neighbors from the student's own history: four are cage
problems answered incorrectly, at weights $0.19$, $0.13$, $0.11$, and
$0.11$, while the only correct neighbor, the same structure disguised
with monsters in place of animals, draws $0.04$. The similarity summary
reads $0.34$ against the student's $0.66$ overall correctness: the
addressed evidence diverges from the student's record in the
anticipated direction, with the wording match concentrating on failures
the student has since learned from. The evidence-volume features name
the cause, calling the addressing diffuse at entropy $3.2$. Weighing
this summary beside the behavioral state, the layer returns
$\hat{y}=0.48$; the student answers correctly.

The miss admits two readings: either the dominant neighbors carry
genuine signal---this student failed exactly this problem family---and
the model rightly lands near the boundary rather than on either side,
or the addressing is too diffuse to bear the weight the layer gave it,
in which case an evidence-aware discount would have moved the
prediction toward the correct side.

\section{Conclusion}
\label{sec:conclusion}

We studied knowledge tracing on a new platform with no response logs
and no shared students, under access rules that admit the target
domain only through its public question texts. The central finding is
a division of labor: correctness dynamics carry the transfer, while
question semantics, segmented into near-disjoint embedding spaces by
domain, serve only to address a student's own history. \textbf{TIZSKT}
implements this division by concentrating trainable capacity in one
behavioral encoder and leaving every use of text parameter-free, and
it reaches the best zero-shot accuracy on every domain.

\balance
\bibliographystyle{ACM-Reference-Format}
\bibliography{references}

\end{document}